\documentclass[11pt]{article}
\usepackage{acl2014}
\usepackage{times}
\usepackage{url}
\usepackage{latexsym}
\usepackage{amssymb}
\usepackage{graphicx}
\usepackage{amsmath}
\usepackage[font=footnotesize]{caption}
\usepackage{multirow}
\usepackage[belowskip=-14pt, aboveskip=2pt]{caption}
\usepackage{balance}
\usepackage{subcaption}

\title{Temporal Analysis of Language through Neural Language Models}
\vspace{5mm}
\author{Yoon Kim$^*$ \hspace{2mm} Yi-I Chiu$^*$ \hspace{2mm} Kentaro Hanaki$^*$ \hspace{2mm}  Darshan Hegde$^*$ \hspace{2mm} Slav Petrov$^\diamond$ \\ \\
  $^*$New York University, New York \\
  $^\diamond$Google Inc., New York  \\ \\
  {\tt \{yhk255, yic211, kh1615, dh1806\}@nyu.edu} \\
  {\tt slav@google.com}}
\date{}

\begin{document}
\maketitle
\begin{abstract}
We provide a method for automatically detecting change in language across time through a chronologically trained neural language model. We train the model on the Google Books Ngram corpus to obtain word vector representations specific to each year, and identify words that have changed significantly from 1900 to 2009. The model identifies words such as \emph{cell} and \emph{gay} as having changed during that time period. The model simultaneously identifies the specific years during which such words underwent change. \end{abstract}
\section{Introduction}
Language changes across time. Existing words adopt additional senses (\emph{gay}), new words are created (\emph{internet}), and some words `die out' (many irregular verbs, such as \emph{burnt}, are being replaced by their regularized counterparts \cite{Lieberman:2007}). Traditionally, scarcity of digitized historical corpora has prevented applications of contemporary machine learning algorithms---which typically require large amounts of data---in such temporal analyses. Publication of the Google Books Ngram corpus in 2009, however, has contributed to an increased interest in \emph{culturomics}, wherein researchers analyze changes in human culture through digitized texts \cite{Michel:2011}. 

Developing computational methods for detecting and quantifying change in language is of interest to theoretical linguists as well as NLP researchers working with diachronic corpora. Methods employed in previous work have been varied, from analyses of word frequencies to more involved techniques (Guolordava et al. \shortcite{Gulordava:2011}; Mihalcea and Nataste \shortcite{Mihalcea:2012}). In our framework, we train a Neural Language Model (NLM) on yearly corpora to obtain word vectors for each year from 1900 to 2009.  We chronologically train the model by initializing word vectors for subsequent years with the word vectors obtained from previous years.

We compare the cosine similarity of the word vectors for same words in different years to identify words that have moved significantly in the vector space during that time period. Our model identifies words such as \emph{cell} and \emph{gay} as having changed between 1900--2009. The model additionally identifies words whose change is more subtle. We also analyze the yearly movement of words across the vector space to identify the specific periods during which they changed. The trained word vectors are publicly available.\footnote{http://www.yoon.io}

\section{Related Work}
Previously, researchers have computationally investigated diachronic language change in various ways. Mihalcea and Nastase \shortcite{Mihalcea:2012} take a supervised learning approach and predict the time period to which a word belongs given its surrounding context. Sagi et al. \shortcite{Sagi:2009} use a variation of Latent Semantic Analysis to identify semantic change of specific words from early to modern English. Wijaya and Yeniterzi \shortcite{Wijaya:2011} utilize a Topics-over-Time model and K-means clustering to identify periods during which selected words move from one topic/cluster to another. They correlate their findings with the underlying historical events during that time. Gulordava and Baroni \shortcite{Gulordava:2011} use co-occurrence counts of words from 1960s and 1990s to detect semantic change. They find that the words identified by the model are consistent with evaluations from human raters. Popescu and Strapparava \shortcite{Popescu:2013} employ statistical tests on frequencies of political, social, and emotional words to identify and characterize epochs.

Our work contributes to the domain in several ways. Whereas previous work has generally involved researchers manually identifying words that have changed (with the exception of Gulordava and Baroni \shortcite{Gulordava:2011}), we are able to automatically identify them. We are additionally able to capture a word's yearly movement and identify periods of rapid change. In contrast to previous work, we simultaneously identify words that have changed and also the specific periods during which they changed.

\section{Neural Language Models}
Similar to traditional language models, NLMs involve predicting a set of future word given some history of previous words. In NLMs however, words are projected from a sparse, 1-of-V encoding (where V is the size of the vocabulary) onto a lower dimensional vector space via a hidden layer. This allows for better representation of semantic properties of words compared to traditional language models (wherein words are represented as indices in a vocabulary set). Thus, words that are semantically close to one another would have word vectors that are likewise `close' (as measured by a distance metric) in the vector space. In fact, Mikolov et al. \shortcite{Mikolov:2013a} report that word vectors obtained through NLMs capture much deeper level of semantic information than had been previously thought. For example, if $x_w$ is the word vector for word $w$, they note that $x_{apple} - x_{apples} \approx x_{car} - x_{cars} \approx x_{family} - x_{families}$. That is, the concept of pluralization is learned by the vector representations (see Mikolov et al. \shortcite{Mikolov:2013a} for more examples).

NLMs are but one of many methods to obtain word vectors---other techniques include Latent Semantic Analysis (LSA) \cite{Deerwester:1990}, Latent Dirichlet Allocation (LDA) \cite{Blei:2003}, and variations thereof. And even within NLMs there exist various architectures for learning word vectors (Bengio et al. \shortcite{Bengio:2003}; Mikolov et al. \shortcite{Mikolov:2010}; Collobert et al. \shortcite{Collobert:2011}; Yih et al. \shortcite{Yih:2011}). We utilize an architecture introduced by Mikolov et al. \shortcite{Mikolov:2013b}, called the Skip-gram, which allows for efficient estimation of word vectors from large corpora.

In a Skip-gram model, each word in the corpus is used to predict a window of surrounding words (Figure 1). To ensure that words closer to the current word are given more weight in training, distant words are sampled less frequently.\footnote{Specifically, given a maximum window size of $W$, a random integer $R$ is picked from range [1, $W$] for each training word. The current training word is used to predict $R$ previous and $R$ future words.} Training is done through stochastic gradient descent and backpropagation. The word representations are found in the hidden layer. 
\begin{figure}
\begin{center}
    \includegraphics[scale=0.30]{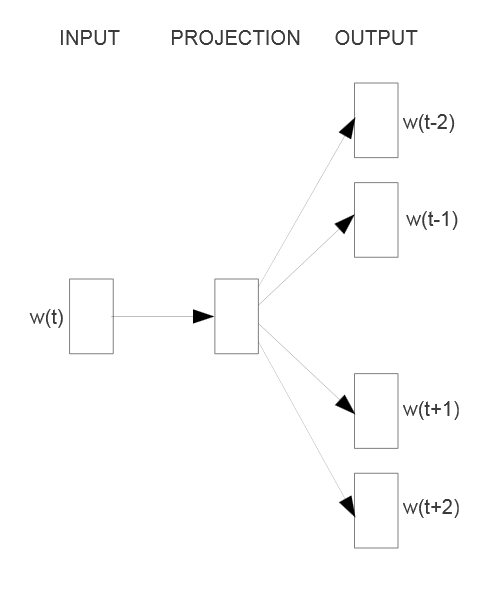}
\caption{Architecture of a Skip-gram model \cite{Mikolov:2013b}.}
\end{center}
\end{figure}
Despite its simplicity---and thus, computational efficiency---compared to other NLMs, Mikolov et al. \shortcite{Mikolov:2013b} note that the Skip-gram is competitive with other vector space models in the Semantic-Syntactic Word Relationship test set when trained on the same data. 
 
\subsection{Training}
The Google Books Ngram corpus contains Ngrams from approximately 8 million books, or 6$\%$ of all books published \cite{Lin:2012}. We sample 10 million 5-grams from the English fiction corpus for every year from 1850--2009. We lower-case all words after sampling and restrict the vocabulary to words that occurred at least 10 times in the 1850--2009 corpus.

For the model, we use a window size of 4 and dimensionality of 200 for the word vectors. Within each year, we iterate over epochs until convergence, where the measure of convergence is defined as the average angular change in word vectors between epochs. That is, if $V(y)$ is the vocabulary set for year $y$, and  $x_w(y,e)$ is the word vector for word $w$ in year $y$ and epoch number $e$, we continue iterating over epochs until,
\begin{equation*}
\frac{1}{|V(y)|}\sum_{w \in V(y)} \arccos{\frac{x_w(y,e)\cdot x_w(y,e-1)}{\|x_w(y,e)\|\|x_w(y,e-1)\|}}
\end{equation*}
is below some threshold. The learning rate is set to 0.01 at the start of each epoch and linearly decreased to 0.0001.

Once the word vectors for year $y$ have converged, we initialize the word vectors for year $y+1$ with the previous year's word vectors and train on the $y+1$ data until convergence. We repeat this process for 1850--2009. Using an open source implementation in the \texttt{gensim} package, training took approximately 4 days on a 2.9 GHz machine.
\section{Results and Discussion}
For the analysis, we treat 1850--1899 as an initialization period and begin our study from 1900.
\subsection{Word Comparisons}
By comparing the cosine similarity between same words across different time periods, we are able to detect words whose usage has changed. We are also able to identify words that did not change. Table 1 has a list of 10 most/least changed words between 1900 and 2009. We note that almost all of the least changed words are function words. For the changed words, many of the identified words agree with intuition (e.g. \emph{gay}, \emph{cell}, \emph{ass}). Others are not so obvious (e.g. \emph{checked}, \emph{headed}, \emph{actually}). To better understand how these words have changed, we look at the composition of their neighboring words for 1900 and 2009 (Table 2).

\begin{table}
\centering
\footnotesize
\begin{tabular}{ |c|c||c|c|} \hline
\multicolumn{2}{|c||}{Most Changed}  &
\multicolumn{2}{|c|}{Least Changed} \\
\hline
Word & Similarity & Word & Similarity \\ 
\hline
\it{checked} & 0.3831 & \it{by} & 0.9331 \\
\it{check} & 0.4073 & \it{than} & 0.9327 \\
\it{gay} & 0.4079 & \it{for} & 0.9313 \\
\it{actually} & 0.4086 & \it{more} & 0.9274 \\
\it{supposed} & 0.4232 & \it{other} & 0.9272 \\
\it{guess} & 0.4233 & \it{an} & 0.9268 \\
\it{cell} & 0.4413 & \it{own} & 0.9259 \\
\it{headed} & 0.4453 & \it{with} & 0.9257 \\
\it{ass} & 0.4549 & \it{down} & 0.9252 \\
\it{mail} & 0.4573 & \it{very} & 0.9239 \\ 
\hline
\end{tabular}
\caption{Top 10 most/least changed words from 1900--2009, based on cosine similarity of words in 2009 against their 1900 counterparts. Infrequent words (words that occurred less than 500 times) are omitted.}
\end{table}
\begin{table}
\centering
\footnotesize
\begin{tabular}{|c|c|c|} \hline
\multirow{2}{*}{Word} & \multicolumn{2}{|c|}{Neighboring Words in} \\ \cline{2-3} & 1900 & 2009 \\ \hline \hline
\multirow{3}{*}{\it{gay}} & \it{cheerful} & \it{lesbian} \\ & \it{pleasant} & \it{bisexual} \\ & \it{brilliant} & \it{lesbians} \\ \hline
\multirow{3}{*}{\it{cell}} & \it{closet} & \it{phone} \\ & \it{dungeon} & \it{cordless} \\ & \it{tent} & \it{cellular} \\ \hline
\multirow{3}{*}{\it{checked}} & \it{checking} & \it{checking} \\ & \it{recollecting} & \it{consulted} \\ & \it{straightened} & \it{check} \\ \hline
\multirow{3}{*}{\it{headed}} & \it{haired} & \it{heading} \\ & \it{faced} & \it{sprinted} \\ & \it{skinned} & \it{marched} \\ \hline
\multirow{3}{*}{\it{actually}} & \it{evidently} & \it{really} \\ & \it{accidentally} & \it{obviously} \\ & \it{already} & \it{nonetheless} \\ \hline
\end{tabular}
\caption{Top 3 neighboring words (based on cosine similarity) specific to each time period for the words identified as having changed.}
\end{table}
As a further check, we search Google Books for sentences that contain the above words. Below are some example sentences from 1900 and 2009 with the word \emph{checked}:
\\
 
\begin{small}
\hspace{-4mm}1900: ``However, he \emph{checked} himself in time, saying ---'' \\
1900: ``She was about to say something further, but she \emph{checked} herself.'' \\
2009: ``He'd \emph{checked} his facts on a notepad from his back pocket.'' \\
2009: ``I \emph{checked} out the house before I let them go inside.''
\end{small}
\\

At the risk of oversimplifying, the resulting sentences indicate that in the past, \emph{checked} was more frequently used with the meaning \emph{``to hold in restraint''}, whereas now, it is more frequently used with the meaning \emph{``to verify by consulting an authority''} or \emph{``to inspect so as to determine accuracy''}. Given that \emph{check} is a highly polysemous word, this seems to be a case in which the popularity of a word's sense changed over time.

Conducting a similar exercise for \emph{actually}, we obtain the following sentences:
\\

\begin{small}
\hspace{-4mm}1900: ``But if ever he \emph{actually} came into property, she must recognize the change in his position.'' \\
1900: ``Whenever a young gentleman was not \emph{actually} engaged with his knife and fork or spoon ---'' \\
2009: ``I can't believe he \emph{actually} did that!'' \\
2009: ``Our date was \emph{actually} one of the most fun and creative ones I had in years.'' \\
\end{small}
\\
Like the above, this seems to be a case in which the popularity of a word's sense changed over time (from \emph{``to refer to what is true or real''} to \emph{``to express wonder or surprise''}).

\subsection{Periods of Change}
As we chronologically train the model year-by-year, we can plot the time series of a word's distance to its neighboring words (from different years) to detect periods of change. Figure 2 (above) has such a plot for the word \emph{cell} compared to its early neighbors, \emph{closet} and \emph{dungeon}, and the more recent neighbors, \emph{phone} and \emph{cordless}. Figure 2 (below) has a similar plot for \emph{gay}.

\begin{figure}
\centering
\begin{tabular}{c}
    \includegraphics[scale=0.35]{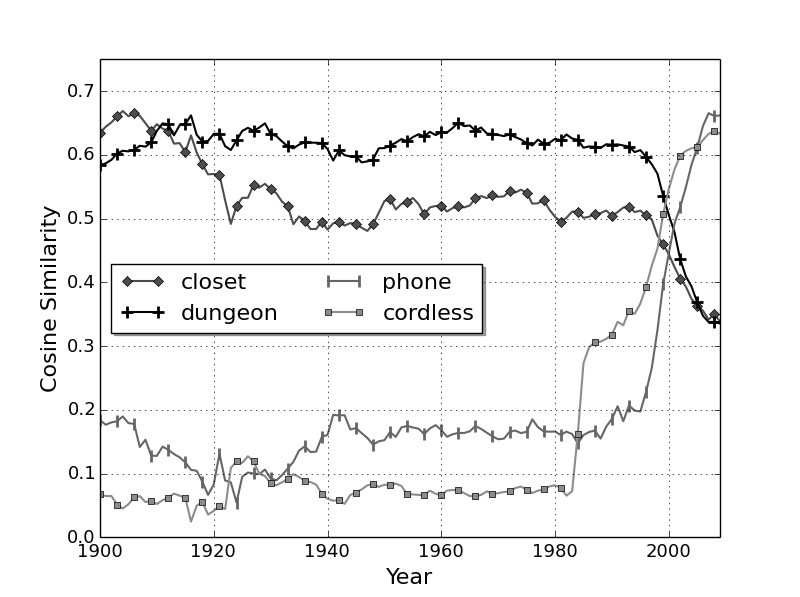} \\
    \includegraphics[scale=0.35]{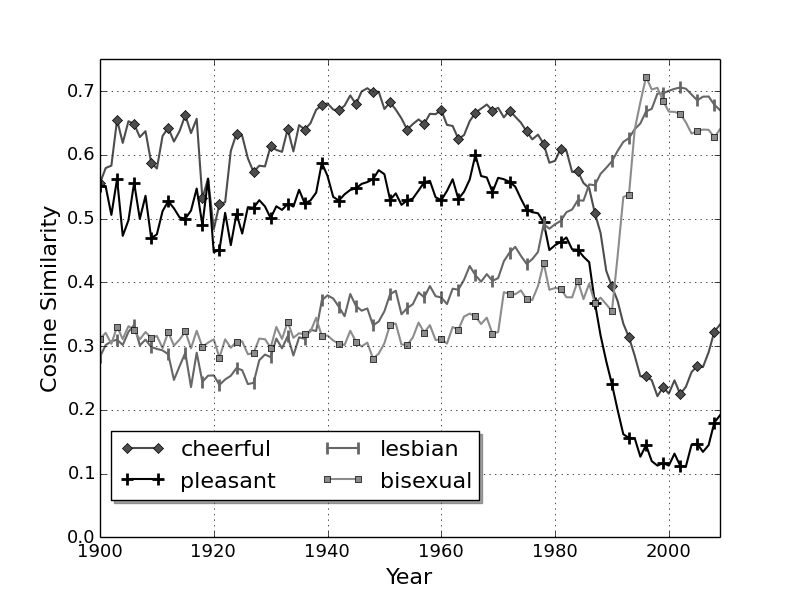}
\end{tabular}
\caption{(Above) Time trend of the cosine similarity between \emph{cell} and its neighboring words in 1900 (\emph{closet}, \emph{dungeon}) and 2009 (\emph{phone}, \emph{cordless}). (Below) Similar plot of \emph{gay} and its neighboring words in 1900 (\emph{cheerful}, \emph{pleasant}) and 2009 (\emph{lesbian}, \emph{bisexual}).}
\end{figure}

Such plots allow us to identify a word's period of change relative to its neighboring words, and thus provide context as to how it evolved. This may be of use to researchers interested in understanding (say) when \emph{gay} started being used as a synonym for \emph{homosexual}. We can also identify periods of change independent of neighboring words by analyzing the cosine similarity of a word against itself from a reference year (Figure 3). As some of the change is due to sampling and random drift, we additionally plot the average cosine similarity of all words against their reference points in Figure 3. This allows us to detect whether a word's change during a given period is greater (or less) than would be expected from chance.
\begin{figure}
\begin{center}
    \includegraphics[scale=0.35]{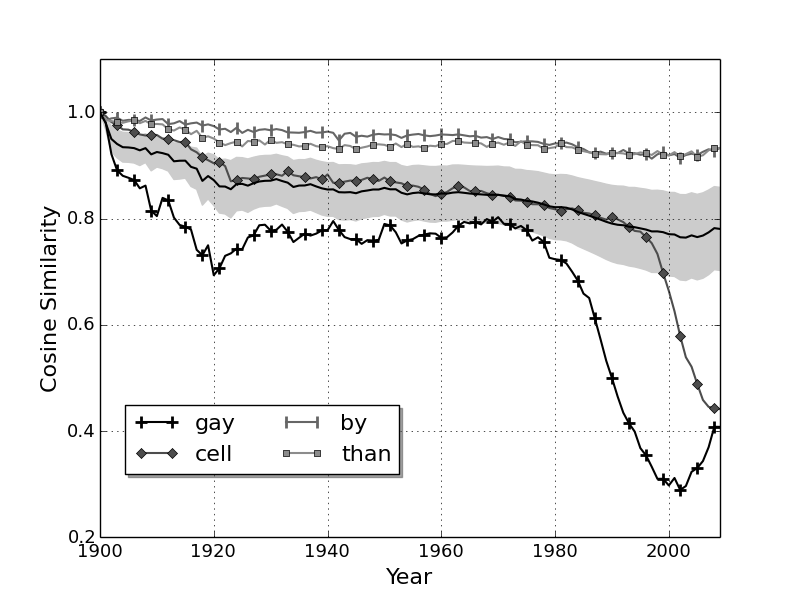}
\caption{Plot of the cosine similarity of changed (\emph{gay}, \emph{cell}) and unchanged (\emph{by}, \emph{than}) words against their 1900 starting points. Middle line is the average cosine similarity of all words against their starting points in 1900. Shaded region corresponds to one standard deviation of errors.}
\end{center}
\end{figure}
We note that for \emph{cell}, the identified period of change (1985--2009) coincides with the introduction---and subsequent adoption---of the cell phone by the general public.\footnote{http://library.thinkquest.org/04oct/02001/origin.htm} Likewise, the period of change for \emph{gay} agrees with the gay movement which began around the 1970s \cite{Wijaya:2011}.

\subsection{Limitations}
In the present work, identification of a changed word is conditioned on its occurring often enough in the study period. If a word's usage decreased dramatically (or stopped being used altogether), its word vector will have remained the same and hence it will not show up as having changed. One way to overcome this may be to combine the cosine distance and the frequency to define a new metric that measures how a word's usage has changed. 

\section{Conclusions and Future Work}
In this paper we provided a method for analyzing change in the written language across time through word vectors obtained from a chronologically trained neural language model. Extending previous work, we are able to not only automatically identify words that have changed but also the periods during which they changed. While we have not extensively looked for connections between periods identified by the model and real historical events, they are nevertheless apparent.

An interesting direction of research could involve analysis and characterization of the different types of change. With a few exceptions, we have been deliberately general in our analysis by saying that a word's \emph{usage} has changed. We have avoided inferring the \emph{type} of change (e.g. semantic vs syntactic, broadening vs narrowing, pejoration vs amelioration). It may be the case that words that undergo (say) a broadening in senses exhibit regularities in how they move about the vector space, allowing researchers to characterize the type of change that occurred.

\balance

\end{document}